\documentclass[conference,10pt]{IEEEtran}
\usepackage{subfigure}
\usepackage{graphicx}
\usepackage{caption}
\usepackage{color}
\usepackage{dsfont}
\usepackage{float}
\usepackage{amsmath}

\title{Context Exploitation using Hierarchical Bayesian Models}
\author{
  \IEEEauthorblockN{Christopher A. George, Pranab Banerjee, Kendra E. Moore}
  \IEEEauthorblockA{Boston Fusion Corp., 70 Westview Street Ste. 100, Lexington, MA 02421 \\ alex.george@bostonfusion.com}
}

\begin{document}
\maketitle

\begin{abstract}
We consider the problem of how to improve automatic target recognition by fusing the na\"ive sensor-level classification decisions with ``intuition,'' or context, in a mathematically principled way. This is a general approach that is compatible with many definitions of context, but for specificity, we consider context as co-occurrence in imagery. In particular, we consider images that contain multiple objects identified at various confidence levels. We learn the patterns of co-occurrence in each context, then use these patterns as hyper-parameters for a Hierarchical Bayesian Model. The result is that low-confidence sensor classification decisions can be dramatically improved by fusing those readings with context. We further use hyperpriors to address the case where multiple contexts may be appropriate. We also consider the Bayesian Network, an alternative to the Hierarchical Bayesian Model, which is computationally more efficient but assumes that context and sensor readings are uncorrelated.
\end{abstract}

\section{Introduction}
The kill chain, identifying and destroying a target, can be factorized into four steps: (1) obtaining sensor readings, (2) classifying the object from the sensor readings (including sensor fusion when multiple sensors are employed), (3) determining whether the sensor reading is reliable, and (4) making a decision to fire or to hold fire. 

Over the past 50 years, tremendous work has been placed into automating the sensor-exploitation-and-fusion step; this work has produced strong capabilities in the challenging problem of automatic target recognition (ATR). A true data-to-decision framework, however, could strengthen ATR by automating some aspects of the operator review step as well. In particular, one of the tasks the operator performs is to review whether each classification decision seems reasonable in view of its context. For example, consider an operator reviewing a sensor reporting 80\% confidence of Russian Bombers over (a) Kiev and (b) New York City. The operator has three options: (1) na\"ively accepting these sensor readings as received (in this case, reporting an 80\% chance of bombers over New York), (2) rejecting the counter-intuitive sensor readings \textit{pro forma} (this defeats the purpose of using the sensor at all), or (3) manually correcting the sensor readings, for example to 90\% and 0.01\%, respectively. Of these, option (3) is perhaps the most agreeable; however, there is a need for a principled way to perform these corrections.

\begin{figure}
\centering
\subfigure[]{\includegraphics[scale=0.57]{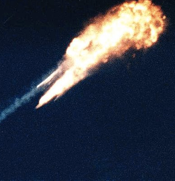}}
\subfigure[]{\includegraphics[scale=0.185]{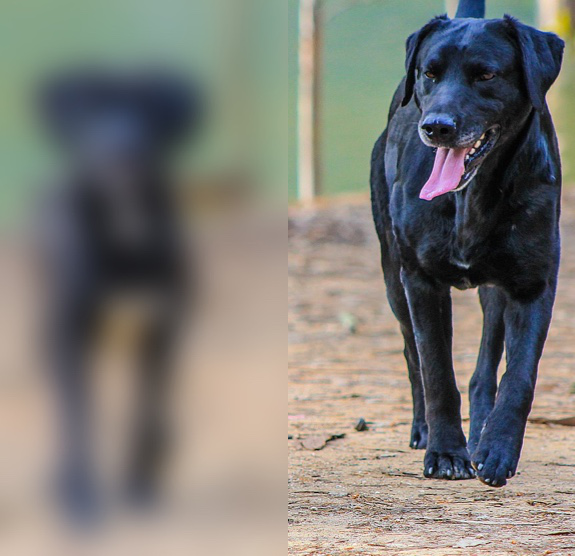}}
\caption{Sample Raw Data}
\label{fig:dogs}
\end{figure}

In this work we consider a general solution to this challenge. For concreteness, we consider the case of imagery datasets (such as those in Figure \ref{fig:dogs}), in which context is defined as knowledge of co-occurrence patterns among object classes. Both images in Figure \ref{fig:dogs} contain two objects, one of which the sensor is able to resolve with 99\% confidence (``ACME warhead'' and ``dog'', respectively); the other is too blurred to be resolved. For this blurry object, we postulate that the sensor reports equal probability for each of three different possible object classes. We will show how context can enhance our classification decision for the blurry object in both cases.

\section{Model}
\subsection{Problem Formulation}
We begin by formulating the problem. We have an image, I, with $N \in \mathds{N}$ objects, each of which represents exactly one of  $M \in \mathds{N}$ object classes, and is associated with one out of at least one possible contexts, $c \in C$. For a military situation, $C$ could be various conflict regions or locations; in civilian photography, these could refer to the topic of a collection (sporting events, architecture, portraits, etc). We further have a set of sensor readings, $X$, such that $|X| = N$ (these sensor readings may be decisions from a single sensor or from a multisensor inference algorithm); sensor readings further have a reported uncertainty level.

Our problem is to quantify the context, $C$, and use the context to correct low-confidence sensor readings. We consider two models for this purpose: a Bayesian Network (BN) and a Hierarchical Bayesian Model (HBM).

\subsection{Bayesian Network}
We construct the Bayesian network in a completely data-driven way. Beginning with labeled images $I_c$ associated with context $c$, we construct a graph for each $c$ where each object class $m$ is a node and the edge weights between $m_A$ and $m_B$ represent the number of $I_c$ in which objects of class $A$ and $B$ co-appear. For convenience, we assign parent-child relationships based on the alphabetical ordering of the node names. 

The result is a Bayesian Network that fully encodes the joint distribution over all object classes. Further, the network is efficient in that only nodes that co-occur are connected (further, the threshold for connectivity can be raised to reduce the number of edges). 

\subsection{Hierarchical Bayesian Model}

For the Hierarchical Bayesian Model \cite{HBMbook}, we begin again with the labeled images, $I_c$, associated with context, $c$; for each $c$, we calculate the $\mu$ (an $M$-dimensional vector) and $\Sigma$ (an $M \times M$ dimensional matrix) that give the frequency of each object class and the correlation matrix among the $M$ object classes, respectively. The values for $\mu$ and $\Sigma$ can be measured quantitatively from labeled training data or provided as estimates. 

\begin{figure}[!t]
\centering
\includegraphics[scale=0.2]{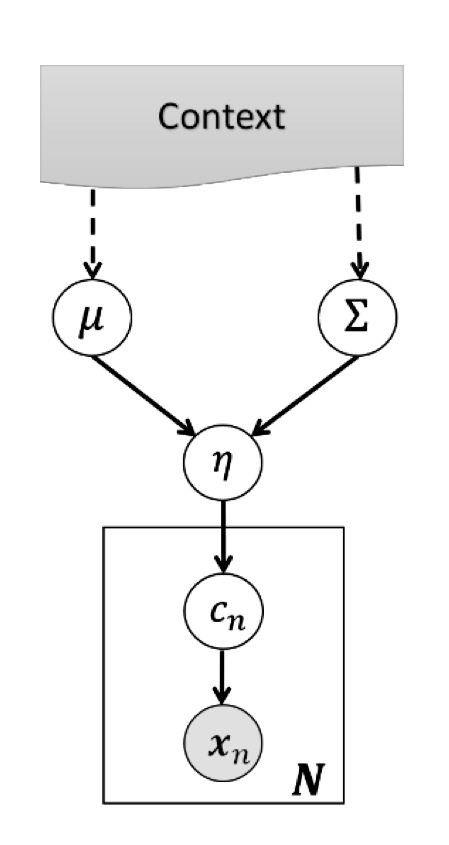}
\caption{Hierarchical Bayesian Model for context-enhanced classification.}
\label{fig:model}
\end{figure}

We now model the sensor readings as having been drawn from a hierarchy of latent probability distributions. This is depicted in plate notation in Figure \ref{fig:model}, in which the shaded plates represent observed variables. In particular, $\mu$ and $\Sigma$ (defined above) are used to define a $N$-dimensional multivariate normal distribution from which some vector, $\eta$, is drawn. We then normalize $\eta'$ to obtain $\eta$ according to the formula:
\begin{equation} \eta_m = N \frac{e^{\eta'}}{\sum_{m \in M} e^{\eta'_m}} \label{eqn:blah} \end{equation}
Each $\eta_i, i \in M$ represents the expected number of times that an object class $i$ will appear in the scene (note that the value of $\eta$ is fully deterministic given the value of $\eta'$; hence, only $\eta$ appears in Figure \ref{fig:model}). 

Given $\eta$, the vectors $c_n$ are then chosen subject to the constraints that $\sum_{n \in N} c_n = \eta$, $\sum_{m \in M} c_{mn} = 1$, and $0 \leq c_{mn} \leq 1$; we interpret $c_{mn}$ as the probability that object $n$ belongs to class $m$. Finally, the sensor readings $x_n$ are drawn from a latent multivariate normal distribution peaked at $c_n$, with a standard deviation given by the sensor's reported uncertainty. This is a key point: if the sensor reports a high degree of certainty, the Hierarchical Bayesian Model will not attempt to second-guess its decision (this is analogous to humans making counter-intuitive classification decisions, i.e., ``I know a tiger when I see one!'').  

To summarize, the HBM is a generative model that generates sensor observations by performing the following steps:
\begin{enumerate}
\item Generate the scene from context via $\eta' \sim N(\mu, \Sigma)$.
\item Normalize $\eta$ according to equation (\ref{eqn:blah}).
\item For each of the $N$ objects, generate the true object class distribution $c_n$ by randomly choosing vectors that sum to $\eta$ and can be interpreted as probabilities (i.e., sum to one, no values below zero or above one).
\item Generate sensor observations using a multivariate normal peaked around $c_n$ with standard deviation given by the sensor's reported uncertainty.
\end{enumerate}
We can now write the joint distribution for the probability distribution of the observations and true values in terms of the hyper-parameters $\mu$ and $\Sigma$. This is given in Equation \ref{bigIntegral}.
\begin{equation}  P(\mathbf{X},\mathbf{C} | \mu, \Sigma) = \int \left( \prod_{n=1}^N p(x_n|c_n)P(c_n|\eta) \right) p(\eta | \mu, \Sigma) d\eta  \label{bigIntegral} \end{equation}

\section{Results}
\subsection{Toy Scenario}
We now define a notional missile-defense scenario under which we can leverage this model. We imagine that there are three types of warheads, which we name ACME, GLOBEX, and TRIOZAP. We further imagine that there are three empires with nuclear technology: the Kingdoms of Ohio, Iowa, and Utah. We imagine that Ohio and Utah commonly use TRIOZAP warheads whereas Iowa rarely use them; further, Ohio often launches ACME and TRIOZAP warheads together, whereas Iowa and Utah rarely launch them together. This scenario is summarized in Table \ref{tab:scenSummary}.

\begin{table}[H]
\centering
\caption{Summary of scenario}
\begin{tabular}{|c|c|c|}
\hline
\textbf{Kingdom} & \multicolumn{2}{|c|}{\textbf{Triozap warheads are...}} \\
\hline
Iowa & Rare & Anti-correlated with ACME \\
\hline
Ohio & Common & Correlated with ACME \\ 
\hline
Utah & Common & Anti-correlated with ACME \\
\hline
\end{tabular}
\label{tab:scenSummary}
\end{table}
Numerically, we assume that the relative frequency of each object class ($\mu$) and the correlation matrix among these object classes ($\Sigma$) are given by:
\begin{eqnarray} 
\mu_{\textrm{Iowa}} = \langle 0.45, 0.45, 0.1 \rangle  \\
\mu_{\textrm{Ohio,Utah}} = \langle 0.1, 0.4, 0.5 \rangle  \\
\Sigma_{\textrm{Ohio}} = \left( \begin{array}{ccc} 1 & 0.9 & -0.65 \\ 0.9 & 1 & -0.9 \\ -0.65 & -0.9 & 1 \end{array} \right) \\
\Sigma_{\textrm{Iowa,Utah}} = \left( \begin{array}{ccc} 1 & 0 & -0.9 \\ 0 & 1 & 0.3 \\ -0.9 & 0.3 & 1 \end{array} \right) 
\end{eqnarray}
We now imagine that an imaging sensor returns Figure \ref{fig:dogs}(a), from which two objects are detected: an ACME warhead at 99\% $\pm$ 1\% confidence, and an unknown warhead that is consistent with ACME, GLOBEX, or TRIOZAP at 33.33\% $\pm$ 30\% confidence each. 

\subsection{HBM Results} 
We attempt to use these hyper-parameters to solve equation \ref{bigIntegral}. This integral turns out to be intractable (this is common for HBMs); however, we can sample the latent true object distribution, $c$, using a sampling technique such as Markov Chain Monte Carlo (MCMC); Table \ref{tab:Results} shows the results.. 
\begin{table}[!h]
\centering
\caption{Results}
\begin{tabular}{|c|c|c|c|}
\hline
\textbf{\#} & \textbf{Context} & \textbf{Sensor Configuration} & \textbf{P(TRIOZAP)} \\
\hline
1 & None & Physical Sensor Alone & 0.334 \\
\hline
2 & Utah & Physical Sensor $+$ context, $\Sigma = 1$ & 0.423 \\
\hline
3 & Utah & Physical Sensor $+$ full context & 0.314 \\
\hline
4 & Iowa & Physical Sensor $+$ full context & 0.155 \\ 
\hline
5 & Ohio & Physical Sensor $+$ full context & 0.667 \\
\hline
\end{tabular}
\label{tab:Results}
\end{table}

These results are aligned with our intuition:
\begin{itemize}
\item Line 1 recapitulates our statement that the physical sensor alone gives a 33.3\% probability to each of the three warhead classes.
\item Line 2  takes into account only the fact that Utah commonly uses TRIOZAP missiles; P(TRIOZAP) increases accordingly.
\item Line 3 takes into account both that Utah commonly uses TRIOZAP missiles and that TRIOZAP missiles are anti-correlated with ACME; these results largely cancel. 
\item Lines 4 \& 5 give the corresponding results for Iowa and Ohio; the strength of the contextual expectation coupled with the very uncertain sensor reading results in a substantial correction magnitude.
\end{itemize}

We pause to take advantage of the opportunity to compare sampling methods as implemented in the Python scripting language (as of early 2017). In particular, we use the PyMC package for Python 2.7 to sample the latent true object distribution via MCMC (``MCMC2''), and the same package for Python 3.6 to sample the latent true object distribution via MCMC (``MCMC3''), the Variational Bayes (VB) technique, and the Hamiltonian Monte Carlo (HMC) as implemented via the No-U-Turn Sampler (NUTS) \cite{NUTS}. In all cases we use the default parameters, as the PyMC package claims to require minimal fine-tuning. We show our results in Table \ref{tab:samplingResults}, in which the algorithm is defined to have converged when no more than 3 of the 20 Geweke Scores are greater than $\pm$ 0.01; this is quite a conservative criterion. 

\begin{table}[!h]
\centering
\caption{Comparison of sampling algorithms}
\begin{tabular}{|c|c|c|c|c|}
\hline
\textbf{Sampling} & \textbf{P(Triozap)} & \textbf{P(Triozap)} & \textbf{Time} & \textbf{Iter. to} \\
\textbf{Method} & \textbf{Ohio} & \textbf{Utah} & \textbf{per iter} & \textbf{Converge} \\
\hline
MCMC2 & 0.667 & 0.155 & 158 $\mu$s & $\sim$20k \\
\hline
MCMC3 & 0.668 & 0.107 & 680 $\mu$s & $\sim$20k \\
\hline
HMC & 0.667 & 0.109 & 23,484 $\mu$s & $\sim$50k \\
\hline
VB & 0.747 & 0.186 & 236 $\mu$s & $\sim$30k \\
\hline
\end{tabular}
\label{tab:samplingResults}
\end{table}
Table \ref{tab:samplingResults} shows that although all four methods give similar results, the VB method is notably different and the HMC technique is notably slow. Further, the interface to the PyMC package for Python 2 is far more user-friendly.  We therefore quote all results using PyMC for Python 2. Though it is unfortunate that these different sampling methods give non-trivially different results, this was a known issue at the time of this work, see \cite{ERROR}.

\subsection{Hyperpriors}
There is one final generalization we can consider: what if we have the image, but do not know the context? In this case, we can simply take the weighted average of the contexts of which we are aware; the weighted coefficients can be added to our model. These weighted coefficients are the hyperpriors.

To test this approach, we consider an extreme case in which we have an image with three ACME warheads and three GLOBEX warheads as well as blurry object. We do not know which context (country) this comes from, but a clever human would suspect Iowa (because Iowa does not use many TRIOZAP warheads). Indeed, a flat hyperprior chooses a 49\% chance of Iowa, a 26\% chance of Utah, and only a 25\% chance of Ohio. Using these hyperpriors, the final probability for the blurry object to be a TRIOZAP warhead is therefore changed from 33.3\% to 36.6\%. These hyperpriors therefore allow us to leverage context even when we do not know which context to leverage.

\subsection{Results on Microsoft's COCO Dataset}
As a real-world demonstration, we use Microsoft's COCO dataset \cite{DBLP:journals/corr/LinMBHPRDZ14}, which contains over 82,000 hand-labeled images in which all instances of 80 categories (e.g., person, car, bus, hot dog) are labeled. These 80 categories are grouped into 12 super-categories (e.g., outdoor, animal, sports). We now apply both the Bayesian Network and the Hierarchical Bayesian Model to this dataset, and use these learned models to resolve the blurry object shown in Figure \ref{fig:dogs}(b). 

\subsubsection{BN}
\begin{figure*}
\centering
\includegraphics[scale=0.6]{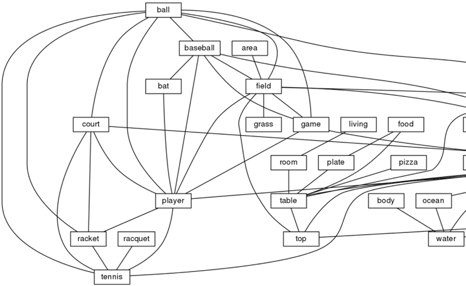}
\caption{Bayesian Network}
\label{fig:BayNet}
\end{figure*}

We use the entire COCO dataset to construct a Bayesian Network as described previously. In particular, we connect two nodes only if their edge weight (number of images in which the object classes co-occur, out of $\sim82,000$ total images) is greater than 1000. Part of the resulting Bayesian Network is depicted in Figure \ref{fig:BayNet}. 

We then use this network to compute our contextual expectation for an image similar to Figure \ref{fig:dogs}(b), in which one figure is clearly a ``street'' and the other figure is blurry. From the Bayesian Network, we can use standard inference techniques to calculate the probabilities for the blurry object as shown in Table \ref{tab:BN}. 
\begin{table}
\centering
\caption{BN Results, COCO context}
\begin{tabular}{|c|c|}
\hline
\textbf{Object} & \textbf{Probability of co-occurrence} \\
\hline
Man & 0.240 \\
\hline 
Road & 0.151 \\
\hline 
Sign & 0.074 \\
\hline 
Sidewalk & 0.051 \\
\hline
Traffic & 0.076 \\
\hline
\end{tabular}
\label{tab:BN}
\end{table}

If the physical sensor provides observations similar to the above (even at low confidence), then the concordance between the sensor reading and the contextual expectation can augment the operator's confidence in the sensor output. If the sensor readings are very dissimilar from Table \ref{tab:BN} (especially if the sensor confidence is low), then the operator should be more cautious of these sensor readings. The BN alone, however, does not provide a principled way to combine these divergent strategies.

\subsubsection{HBM}
As with the BN, we again use the COCO dataset to quantify context; in particular, we calculate $\mu$ and $\Sigma$ from the image co-occurrence. As an additional exercise, we use the 12 super-categories described above as different contexts, and calculate $\mu$ and $\Sigma$ for the 80 categories separately for each of the 12 contexts (by definition, each super-category will be dominated by objects from the categories that make up the super-category; however, other categories will also be represented due to co-occurrence). 

Given these hyper-parameters, we address Figure \ref{fig:dogs}(b), in which one object is clearly a dog, while we posit that the sensor tells us that the blurry object is consistent with another dog, a cat, or a pair of skis. As before, we imagine that all of these hypotheses are equally likely (according to the sensor), with  $33.3\% \pm 30\%$ confidence each. The HBM (trained on all images) substantially decreases the likelihood of skis (skis are rarely photographed at all, and even less commonly with dogs). More specific contexts give even more emphatic results, however: if the photo is evaluated under the ``animal'' context, the ski hypothesis is almost completely rejected; while in the ``sports'' context, the skis become the most likely outcome by far. This is consistent with our intuition: if the photo is labeled ``sports'' (e.g., if it appears in \textit{Sports Illustrated}), then we would expect the image to contain something sport-relevant; this expectation would be very different if the photo appeared in \textit{Animal Planet}. These numerical results are given in Table \ref{tab:dogResults}. 
\begin{table}[!h]
\centering
\caption{HBM Results, COCO context}
\begin{tabular}{|c|c|}
\hline
\textbf{Context} & \textbf{P(Skis)} \\ 
\hline
Sensor Alone & 33.3\% \\
\hline
All Contexts & 21.8\% \\ 
\hline
Animal Context & 0.65\% \\
\hline
Sports Context & 63.67\% \\
\hline
\end{tabular}
\label{tab:dogResults}
\end{table}

\section{Conclusions}
We have designed an HBM and used it to enhance low-confidence sensor readings using contextual information, eliminating the need for \textit{ad hoc} adjustments by the analyst (at least for this type of context). We further showed numerical results on cases in which the context was defined as co-occurrence in imagery. We find that, in this setting, the HBM provides a natural way to numerically fuse the contextual expectation with the sensor readings and sensor uncertainties. We also considered using a Bayesian Network for this purpose; though the Bayesian Network assumes that sensor readings and context are uncorrelated, it is more computationally efficient. Though this setting defines context as co-occurrence only, considering other definitions of context would be a natural extension to this work.

\section*{Acknowledgments} 
This work was supported by the Missile Defense Agency under contract HQ0147-16-C-7602. Approved for Public Release 18-MDA-9664 (30 May 2018).

\bibliographystyle{IEEEtran}
\bibliography{bibliography}

\end{document}